\documentclass[10pt,twocolumn,letterpaper]{article}

\usepackage[pagenumbers]{iccv} % To force page numbers, e.g. for an arXiv version

\definecolor{iccvblue}{rgb}{0.21,0.49,0.74}
\usepackage[pagebackref,breaklinks,colorlinks,allcolors=iccvblue]{hyperref}
\usepackage{amsmath}
\usepackage{amssymb}
\usepackage{mathtools}
\usepackage{xcolor} 
\usepackage{graphicx}
\usepackage{wrapfig}
\usepackage{algorithm}
\usepackage{algorithmic}
\usepackage{booktabs}
\usepackage{multirow}
\usepackage{afterpage}
\usepackage{placeins}

\def\paperID{5593} % *** Enter the Paper ID here
\def\confName{ICCV}
\def\confYear{2025}

\title{TPDiff: Temporal Pyramid Video Diffusion Model}

\author{
Lingmin Ran$^1$ \quad Mike Zheng Shou$^{1,}$\footnotemark[1]\\ \\
$^1$Show Lab, National University of Singapore\\
}

\begin{document}
\maketitle

\renewcommand{\thefootnote}{\fnsymbol{footnote}}
\footnotetext[1]{Corresponding Author.}

\begin{abstract}
The development of video diffusion models unveils a significant challenge: the substantial computational demands. To mitigate this challenge, we note that the reverse process of diffusion exhibits an inherent entropy-reducing nature. Given the inter-frame redundancy in video modality, maintaining full frame rates in high-entropy stages is unnecessary.
Based on this insight, we propose TPDiff, a unified framework to enhance training and inference efficiency.
% for different diffusion forms. 
By dividing diffusion into several stages, our framework progressively increases frame rate along the diffusion process with only the last stage operating on full frame rate, thereby optimizing computational efficiency. 
To train the multi-stage diffusion model, we introduce a dedicated training framework: stage-wise diffusion. By solving the partitioned probability flow ordinary differential equations (ODE) of diffusion under aligned data and noise, our training strategy is applicable to various diffusion forms and further enhances training efficiency.
Comprehensive experimental evaluations validate the generality of our method, 
demonstrating 50\% reduction in training cost and 1.5x improvement in inference efficiency.
% demonstrating 2x faster training and 1.5x faster inference. 
Our project page is: \url{https://showlab.github.io/TPDiff/}
% Our model and code will be released at \lm{github url}.
\end{abstract} 
\section{Introduction}
\label{sec:intro}
With the development of diffusion models, video generation has achieved significant breakthroughs. The most advanced video diffusion models~\cite{sora,kling,moviegen} not only enable individuals to engage in artistic creation but also demonstrate immense potential in other fields like robotics~\cite{Rdt-1b} and virtual reality~\cite{Ldm3d-vr}. Despite the powerful performance of video diffusion models, the complexity of jointly modeling spatial and temporal distribution makes their training costs prohibitively high~\cite{opensora, cogvideox, hunyuanvideo}. 
% Moreover, as video length increases, both training costs and generation time scale accordingly.
Moreover, as the demand for long videos increases, the training and inference costs will continue to scale accordingly.

\begin{figure}[t]
    \centering
    \includegraphics[width=\columnwidth]{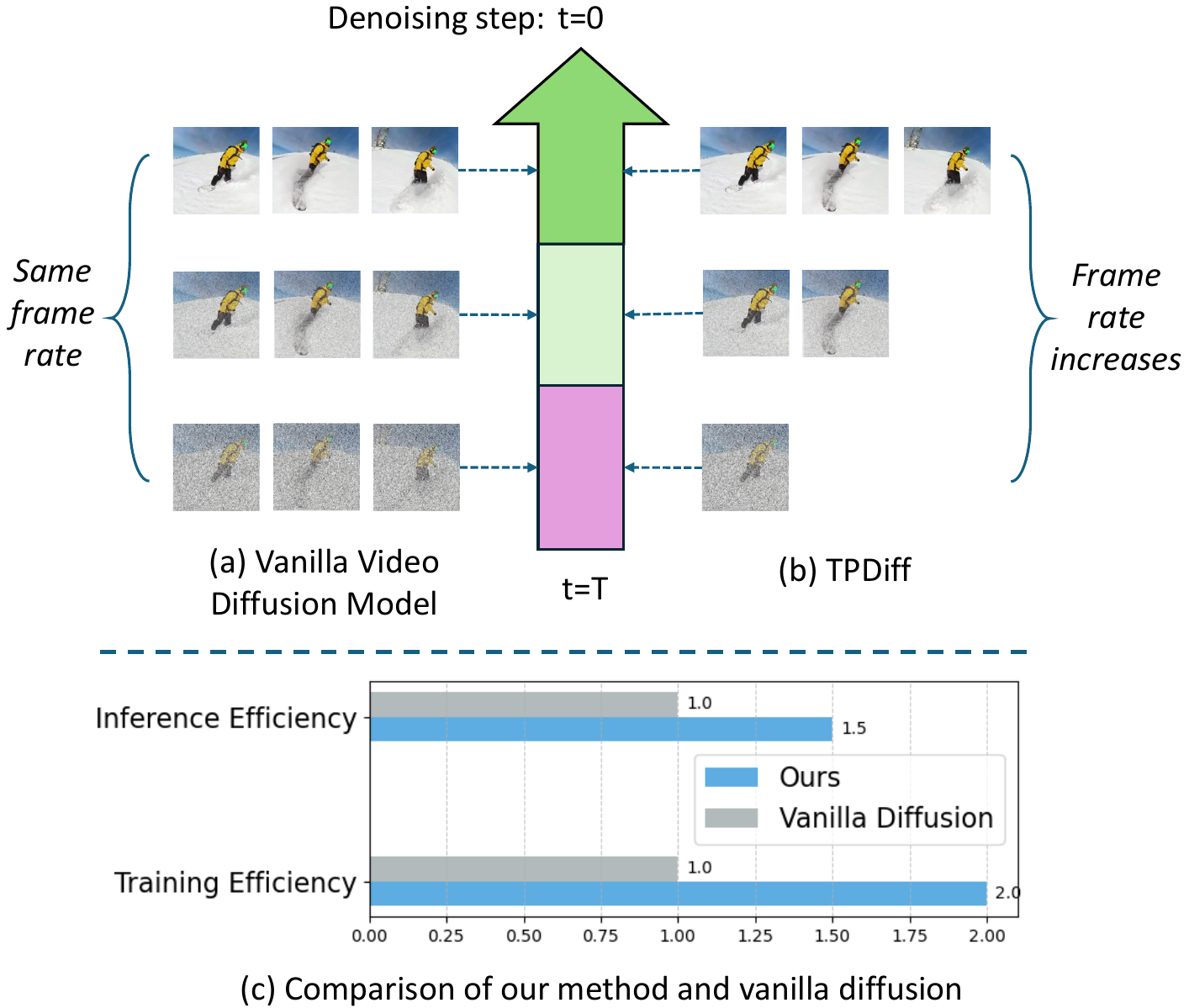}
    % \vspace{-1em}
    % \caption{\textit{\textbf{Overview.}} 
    % \caption{\textit{\textbf{Overview of our method.}}
    % a) Vanilla diffusion models always operate at full frame rates, which is unnecessary and leads to significant computational resource wastage. b) Our method employs progressive frame rates, which utilizes full frame rate only in the final stage, thereby largely optimize computational efficiency in both training and inference.}
    \caption{\textit{\textbf{Overview of our method.}}
    Our method employs progressive frame rates, which utilizes full frame rate only in the final stage as shown in (a) and (b), thereby largely optimizing computational efficiency in both training and inference shown in (c).}
    \vspace{-1em}
    \label{fig:overview}
\end{figure}

To alleviate this problem, researchers propose a series of approaches to increase training and inference efficiency. Show-1~\cite{show-1} and Lavie~\cite{lavie} adopts cascaded framework to model temporal relations at low resolution and apply super-resolution to improve the final video resolution. However, the cascaded structure leads to error accumulation and significantly increases the inference time. 
% Quantization methods~\cite{efficientdm} train diffusion models with lower bit precision while the generated quality inevitably degrades as visual content is sensitive to parameter precision. 
SimDA~\cite{simda} proposes a lightweight model which replaces Attention~\cite{attention} with 3D convolution~\cite{3dconv} to model temporal relationship. Although convolution is computationally efficient, DiT~\cite{dit} demonstrates that attention-based model is scalable and achieves better performance as the volume of data and model parameters increases.
% Among these approaches, the most interesting work is pyramid flow~\cite{PyramidFlow}. 
Recently, ~\cite{PyramidFlow} introduces an interesting work: pyramid flow. 
This method proposes spatial pyramid: it employs low resolution during the early diffusion steps and gradually increases the resolution as the diffusion process proceeds. It avoids the need to always maintain full resolution, significantly reducing computational costs. 

However, pyramid flow has several problems: 1) It only demonstrates its effectiveness under flow matching~\cite{FM} and does not explore its applicability to other diffusion forms like denoising diffusion implicit models~(DDIM)~\cite{ddim}. 
%Additionally, whether similar property exist in the temporal dimension remains unexplored.
2) It formulates video generation in an auto-regressive manner which significantly reduces inference speed. 3) The feasibility of modeling temporal relationship in a pyramid-like structure remains unexplored.
% To accelerate inference process, we will generate all frames in parallel.

To solve the problems, \textit{we propose TPDiff, a general framework to accelerate training and inference speed}. 
% Inspired by previous works~\cite{PyramidFlow, ddim}, 
% in the early timesteps of diffusion, the noisy latents contain limited informative content and the temporal relations between frames are weak. Maintaining full frame rate throughout this process is not essential. 
% Previous works demonstrate that latents in the early timesteps contain limited informative content and the temporal relations between frames are weak, which makes maintaining full frame rate throughout this process unnecessary.
Our method is inspired by the fact that video is a highly redundant modality~\cite{videocompress}, as consecutive frames often contain minimal variations.
Additionally, in a typical diffusion process, latents in the early timesteps contain limited informative content and the temporal relations between frames are weak, which makes maintaining full frame rate throughout this process unnecessary. 
Based on this insight, we propose temporal pyramid: 1) In our method, the frame rate progressively increases as the diffusion process proceeds as shown in 
Fig.~\ref{fig:overview}.
% Fig.~\ref{fig:pipeline}
 Unlike previous works~\cite{lavie, show-1} require an additional temporal interpolation network, we adopt a single model to handle different frame rates. To achieve this, we divide the diffusion process into multiple stages, with each stage operating at different frame rate. 2) To train the temporal pyramid model, we solve the partitioned probability flow ordinary differential equations (ODE)~\cite{ScoreMatching, dpmsolver} by leveraging data-noise alignment and reach a unified solution for various types of diffusion. 3) Our experiments show our method is generalizable to different diffusion forms, including flow matching and DDIM, achieving 2x faster training and 1.5x faster inference compared to vanilla diffusion models. 
% In conclusion, \lm{method name} is a acceleration framework that seamlessly integrates with various video generation frameworks without requiring significant modifications to the training paradigm. 
The core contributions of this paper are summarized as follows:

\begin{itemize}
    \item We introduce temporal pyramid video diffusion model, a generalizable framework aiming at enhancing the efficiency of training and inference for video diffusion models. By employing progressive frame rates across different stages of the diffusion process, the framework achieves substantial reductions in computational cost.
    \item We design a dedicated training framework: stage-wise diffusion. We solve the decomposed probability flow ODE by aligning noise and data. The solution is applicable to different diffusion forms, enabling flexible and seamless extension to various video generation frameworks.
    \item Our experiments demonstrate that the proposed method can be applied across various diffusion frameworks, achieving performance improvement, 2x faster training and 1.5x faster inference.
\end{itemize}
\section{Related works}
\label{sec:related}
\textbf{Generative Video Models}. The field of video generation has witnessed significant progress recently due to the advancement of diffusion models~\cite{Diffusion, DiffusionModel}
% and the large-scale text-video data set~\cite{panda70m}. 
These models generate videos from text descriptions or images. 
% Imagen Video~\cite{Imagen} utilizes a cascading structure for high-resolution text-to-video generation, while the Video Diffusion Model\cite{videodiffusion} expands the standard image architecture to accommodate video data and trains both images and videos together. 
Most methods develop video models based on powerful text-to-image models like Stable Diffusion ~\cite{LDM}, adding extra layers to capture cross-frame motion and ensure consistency. Among these, 
Tune-A-Video~\cite{tuneavideo} employs a causal attention module and limits training module to reduce computational costs. AnimateDiff~\cite{animatediff} utilizes a plug-and-play temporal module to enable video generation on personalized image models~\cite{sd1.5}. 
% Other text-to-video works include marrying latent and pixel space~\cite{show-1} and cascaded genration~\cite{lavie}.
Recently, DiT models~\cite{latte, opensora} pushes the boundaries of video generation. Commercial products~\cite{sora,kling,luma} and open-source works~\cite{hunyuanvideo, cogvideox,opensora} demonstrate remarkable performance by scaling up DiT pretraining. Although DiT achieves significant performance improvements, its training cost escalates to an unaffordable level, hindering the development of video generation.
% Although DiT achieves significant performance improvements, the computational resources required for pretraining increase substantially. 
% Pyramid Flow~\cite{PyramidFlow} introduces a spatial pyramid architecture to reduce pretraining cost. However, its effectiveness has only been validated in flow matching. In our work, we introduce a novel temporal pyramid architecture that not only enhance training and inference efficiency but also demonstrates broad applicability across various diffusion frameworks and parameterizations.

\textbf{Temporal Pyramid}. The complex temporal structure of videos raises a challenge for generation and understanding. 
% In recent years, researchers begin to explore this direction in video understanding. 
SlowFast~\cite{slowfast} simplifies video understanding by utilizing an input-level frame pyramid, where frames at different levels are sampled at varying rates. Each level is independently processed by a separate network, with their mid-level features interactively fused. This combination of the frame pyramid enables SlowFast to efficiently manage the variability of visual tempos. Similarly, DTPN~\cite{DTPN} employs different frame-per-second (FPS) sampling to construct a pyramidal representation for videos of arbitrary length. Temporal pyramid network~\cite{tpn} leverages the feature hierarchy to handle the variance of temporal information. It avoids to learn visual tempos inside a single network, and only need
frames sampled at a single rate at the input-level. Although the effectiveness of temporal pyramid have been validated in video understanding, its application in generation remains under-explored.

\begin{figure*}[t]
    \centering
    \includegraphics[width= 0.95\textwidth]{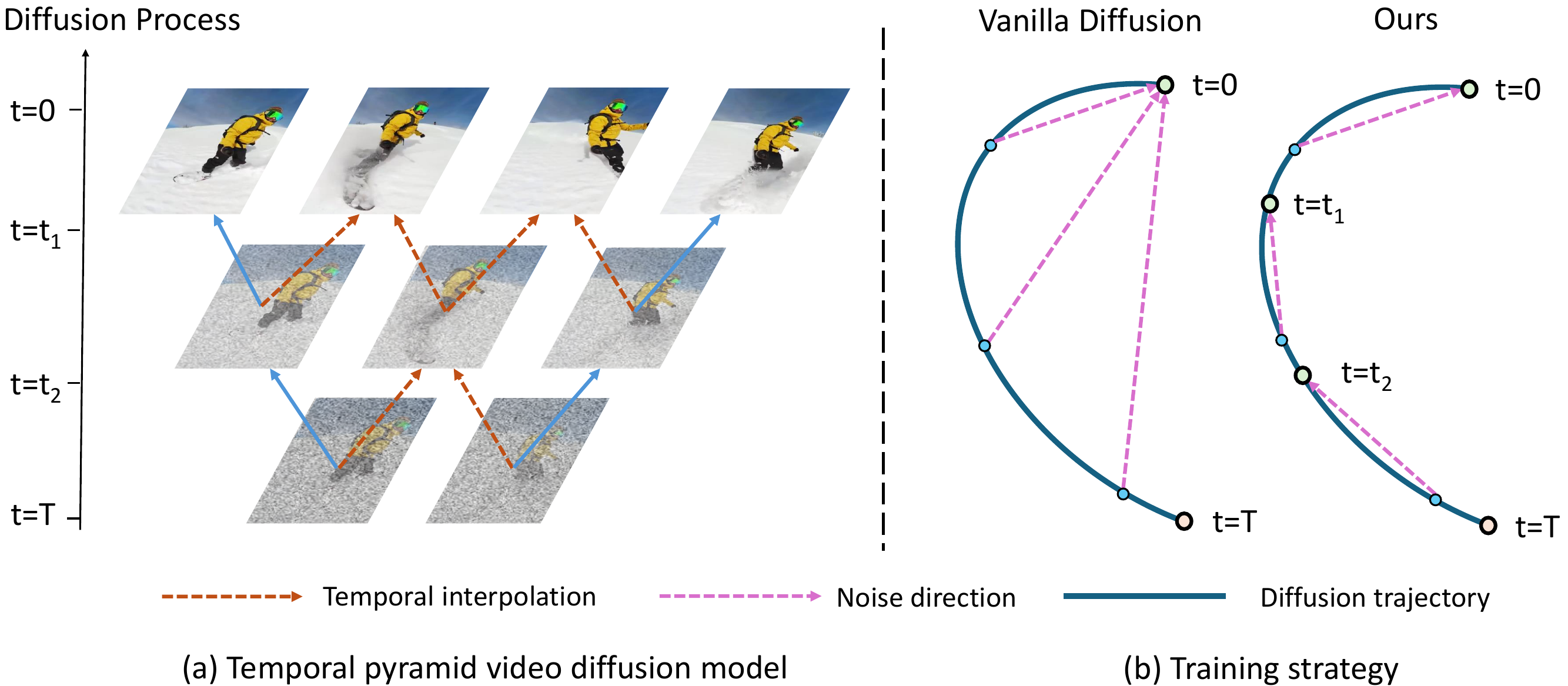}
    % \caption{\textit{\textbf{Method Overview}}. a) Traditional diffusion models consistently operate at full frame rates, which is unnecessary and leads to significant computational resource wastage. b)Our method employs progressive frame rates, namely temporal pyramid, which utilizes full frame rates only in the final stage. Within each stage, new frames are initially interpolated from existing frames at the.}
    \vspace{-0.5em}
    \caption{\textit{\textbf{Methodology}}. a) Pipeline of temporal pyramid video diffusion model. We divide diffusion process into multiple stages with increasing frame rate. In each stage, new frames are initially temporally interpolated from existing frames. b) Our training strategy: stage-wise diffusion. In vanilla diffusion models, the noise direction along the ODE path points toward the real data distribution. In stage-wise diffusion, the noise direction is oriented to the end point of the current stage. }
    \vspace{-1em}
    \label{fig:pipeline}
\end{figure*}

\section{Method}
\label{sec:method}
\subsection{Preliminary}
\label{sec:preliminary}

\paragraph{Denoising Diffusion Implicit Models} DDIM~\cite{ddim} extends DDPMs~\cite{ddpm} by operating in the latent space. Similar to DDPM, in the forward process, DDIM transforms real data $x_0$ into a series of intermediate sample $x_t$, and eventually the Gaussian noise $\epsilon\sim N(0,I)$ according to noise schedule $\overline{\alpha}_t$:
\begin{equation}
\label{equa:ddim_forward}
x_t = \sqrt{\overline{\alpha_t}}x_0 + \sqrt{1-\overline{\alpha}_t}\epsilon, \epsilon\sim N(0,I),
\end{equation}
where $t\sim [1,T]$ and $T$ denotes the total timesteps. After adding noise to the latent, we usually train a neural network $\epsilon_\theta$ to predict the added noise. Formally, $\epsilon_\theta$ is trained using following objecive:
\begin{equation}
\label{equa:ddim_loss}
\min _\theta E_{x_t, \epsilon \sim N(0, I), t \sim \text { Uniform }(1, T)}\left\|\epsilon-\epsilon_\theta\left(x_t, t\right)\right\|_2^2.
\end{equation}
Given a pretrained diffusion model $\epsilon_\theta$, one can generate new data by solving the corresponding probability flow ODE~\cite{ScoreMatching}. DDIM is essentially a first-order ODE solver, which formulates a denoising process to generate $x_{t-1}$ from a sample $x_t$ via: 
\begin{equation}
    x_{t-1} = \alpha_{t-1} \left( \frac{x_t - \sqrt{1-\alpha_t}\epsilon_\theta(x_t, t)}{\alpha_t} \right) + \sqrt{1-\alpha_t}\epsilon_\theta(x_t, t),
\end{equation}
where $\alpha_t=\frac{\overline{\alpha}_t}{\overline{\alpha}_{t-1}}$.

\paragraph{Flow Matching}
Flow-based generative models aim to learn a velocity field $v_\theta$ that transports Gaussian noise $\epsilon\sim N(0,I)$ to the distribution of real data $x_0$. Flow matching~\cite{FM} adopts linear interpolation between noise $\epsilon$ and data $x_0$:
\begin{equation}
\label{equa:fm_forward}
    x_t = (1-t)x_0+t\epsilon, \epsilon\sim N(0,I).
\end{equation}
It trains a neural network $\epsilon_\theta$ to match the velocity field and then solves the ODE for a given boundary condition $\epsilon$ to obtain the flow. The flow matching loss function is as follows:
\begin{equation}
\label{fm_loss}
    \min _\theta E_{x, \epsilon \sim N(0, I), t \sim \text { Uniform }(1, T)}\left\|(\epsilon-x_0)-v_\theta\left(x_t, t\right)\right\|_2^2.
\end{equation}

% Recognizing different diffusion process exhibit a common forward process as shown in Equation~\ref{equa:ddim_forward} and Equation~\ref{equa:fm_forward}, we present a unified diffusion formulation:
% \begin{equation}
% \label{equa:general_diffusion}
%     x_t=\gamma_tx_0+\sigma_t\epsilon,
% \end{equation}
% where the forms of $\gamma_t$ and $\sigma_t$ depend on diffusion framework.

\subsection{Temporal pyramid diffusion}
The core module of existing video diffusion models, attention~\cite{attention}, exhibits quadratic complexity with respect to sequence length.
% , resulting in significantly increased computational cost in long-sequence scenarios such as video generation. 
Our goal is to reduce the sequence length in video generation and decrease the computational cost. Our method is based on two key insights: 1) There is considerable redundancy between consecutive video frames.
2) the early stages of the diffusion process remain at low signal-to-noise ratio (SNR), resulting in minimal information content. It suggests that operating at full frame rate during these initial timesteps is unnecessary. Based on these insights, we propose temporal pyramid video diffusion as shown in Fig.~\ref{fig:pipeline}. Compared to traditional video diffusion model using fixed frame rate, our framework progressively increases the frame rate 
as the denoising proceeds.
% with the denoising progress.
% As introduced in section~\ref{sec:preliminary}, we can present different diffusion frameworks as a general form: 
% % $x_t=\alpha_tx_0+\sigma_t\epsilon$, where $\alpha_t$ and $\sigma_t$ 
% $x_t=\gamma_tx_0+\sigma_t\epsilon$, where $\gamma_t$ and $\sigma_t$ dynamically adjust the weights of the real data and noise at different timesteps. 

In detail, we divide the diffusion process into multiple stages, each characterized by a distinct frame rate, and employ a single model to learn data distributions across all stages. We create $K$ stages $\left\{ \left[ t_k, t_{k-1} \right) \right\}_{k=K}^1$ where $0<t_1<t_2...<t_{K-1}<T$, $T$ denotes the total timesteps. 
% For each time window, let $[s_k,e_k)$ denote the start and end point, the diffusion process for window $k$ is:
% % \begin{equation}
% %     \hat{x}_t = \alpha_{t'}Down(x_{e_k},2^k)+\sigma_{t'}Up(Down({\epsilon}_{s_k},2^{k+1}))
% % \end{equation}
% \begin{equation}
%     % x_t = \alpha_{t}\hat{x}_{e_k}+\sigma_{t}\hat{x}_{s_k}
%     x_t = \gamma_t\hat{x}_{e_k}+\sigma_{t}\hat{x}_{s_k}
% \end{equation}
% where $e_k<t<s_k$, $\hat{x}_{s_k}$ and $\hat{x}_{e_k}$ are start and end point of stage $k$. 
The frame rate at the $k^{th}$ stage is reduced to $\frac{1}{2^{k-1}}$ of the original one. It ensures that only the last stage operates at full frame rate, thereby optimizing computational efficiency. 
% Now the remaining question is how to train a multi-stage diffusion model in a unified manner.
Despite efficiency, the vanilla diffusion model does not support multi-stage training and inference. Therefore, the remaining challenges are: 1) How to train the multi-stage diffsion model in a unified way, which will be introduced in Section~\ref{sec:training_strategy} and Section~\ref{sec:practical_implementation}, 2) How to perform inference, which will be discussed in Section~\ref{sec:inference}.

\subsection{Training strategy}
\label{sec:training_strategy}
% The objective of training is to transport distributions from $\hat{x}_{s_k}$ to $\hat{x}_{e_k}$ at every stage. 
% In the following, we will introduce a unified
% training framework named stage-wise diffusion.

% \paragraph{Stage-wise Diffusion} 
In stage $k$, we denote $(s_k,e_k)$ as the start and end timestep, $\hat{x}_{s_k}$ and $\hat{x}_{e_k}$ as start and end point. 
The objective of training is to transport distribution of $\hat{x}_{s_k}$ to $\hat{x}_{e_k}$ at every stage. 
% To learn each diffusion segment, 
To achieve the objective,
the key is to obtain stage-wise 1) target, \textit{i.e.} $\epsilon$ in DDIM and $\frac{dx_t}{dt}$ in flow matching, and 2) intermediate latents $x_t$ where $t\in [s_k,e_k)$~\cite{RectifiedDiffusion, Perflow}. In the following, we will introduce a unified
training framework named stage-wise diffusion.

\begin{figure}[t]
    \centering
    \includegraphics[width=\columnwidth]{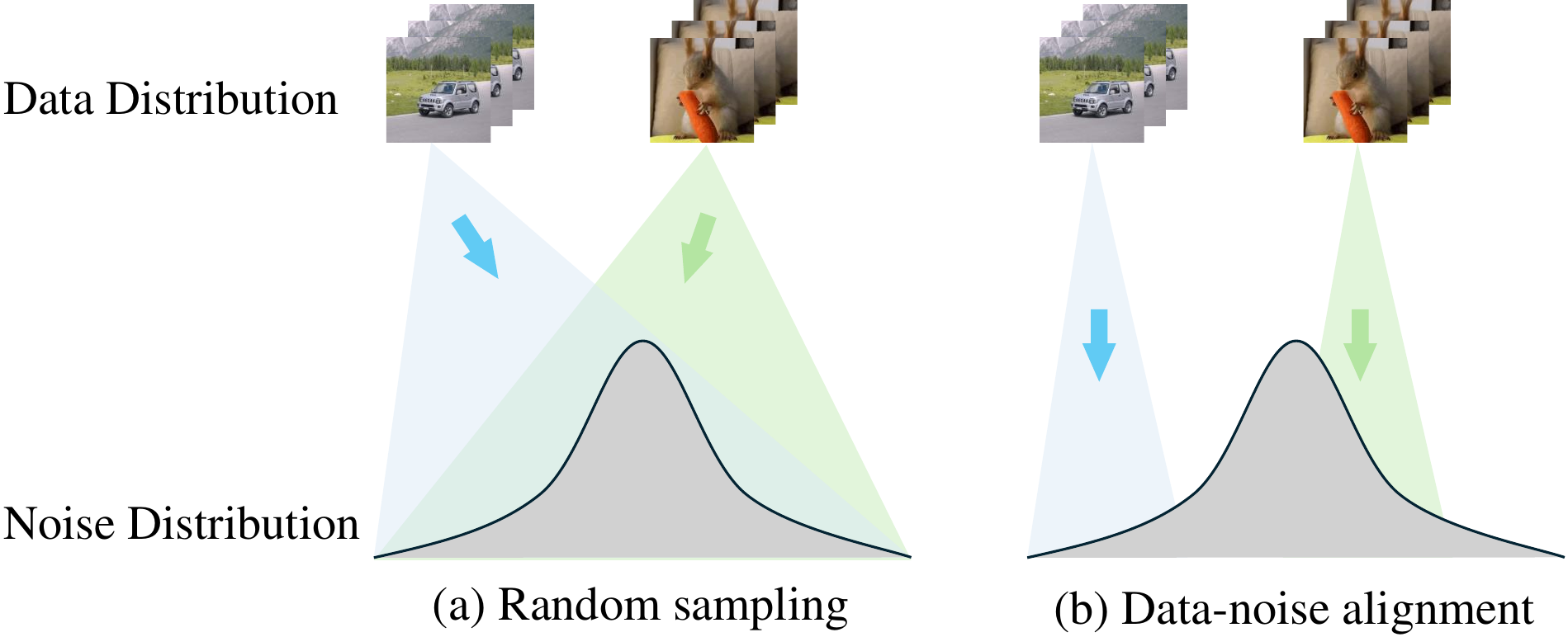}
    % \vspace{-1em}
    \caption{\textit{\textbf{Data-Noise Alignment.}} For every training sample, (a) vanilla diffusion training randomly samples noises across the entire noise distribution, resulting in stochastic ODE path during training. (b) In contrast,  our method samples noises in the closest range, making the ODE path approximately deterministic during training.}
    \label{fig:data_noise_align}
    \vspace{-1em}
\end{figure}

\paragraph{Stage-wise Diffusion} 
To ensure generality, recognizing different diffusion frameworks share a similar formulation as shown in Equation~\ref{equa:ddim_forward} and Equation~\ref{equa:fm_forward}, we present a unified diffusion form:
\begin{equation}
\label{equa:general_diffusion}
    x_t=\gamma_tx_0+\sigma_t\epsilon,
\end{equation}
where the form of $\gamma_t$ and $\sigma_t$ depend on diffusion framework selected. 
Our derivation is based on Equation~\ref{equa:general_diffusion}, without constraining the parameterization of $\gamma_t$ and $\sigma_t$.
% Since each stage has distinctive frame rate, considering stage continuity, 
Considering continuity between stages with distinct frame rates,
we obtain $\hat{x}_{s_k}$ and $\hat{x}_{e_k}$ by:
\begin{equation}
\label{equa:x_s_k}
    % \hat{x}_{s_k}=\alpha_{s_k}Up(Down(x_0,2^{k+1}))+\sigma_{s_k}\epsilon
    \hat{x}_{s_k}=\gamma_{s_k}Up(Down(x_0,2^{k+1}), 2)+\sigma_{s_k}\epsilon,
\end{equation}
\begin{equation}
\label{equa:x_e_k}
    % \hat{x}_{e_k}=\alpha_{e_k}Down(x_0,2^k)+\sigma_{e_k}\epsilon
    \hat{x}_{e_k}=\gamma_{e_k}Down(x_0,2^k)+\sigma_{e_k}\epsilon,
\end{equation}
where $\epsilon\sim N(0,I)$, $Down(\cdot,2^k)$ and $Up(\cdot,2^k)$ are downsampling and upsampling $2^k$ times along temporal axis. We derive the start point of current stage from the end point of preceding stage in Equation~\ref{equa:x_s_k} to bridge adjacent stages, which is crucial for inference and will be introduced in Section~\ref{sec:inference}.
% The objective of multi-stage training is to transport distributions from $\hat{x}_{s_k}$ to $\hat{x}_{e_k}$ at every stage. To learn each diffusion segment, the key is to obtain stage-wise 1) target and 2) intermidiate latents $x_t$~\cite{RectifiedDiffusion, Perflow}. 
% For DDIM diffusion form
% For general diffusion form
% $x_t=\alpha_tx_0+\beta_t\epsilon$,
% $x_t=\gamma_tx_0+\sigma_t\epsilon$,
However, this design also leads to boundary distribution shift and we cannot directly obtain training target from Equation~\ref{equa:x_s_k} and Equation~\ref{equa:x_e_k}. Instead, we should compute added noise in every stage with boundary condition $\hat{x}_{s_k}$ and $\hat{x}_{e_k}$. Fortunately, DPM-Solver~\cite{dpmsolver} derives the relationship between any two points, $x_s$ and $x_e$ on diffusion ODE path and this relationship can also be applied to any stage in our method. Accordingly, in stage $k$, by replacing $x_s$ with $\hat{x}_{s_k}$ and $x_e$ with $x_t$, we can express intermediate latent $x_t$ as a function of $\hat{x}_{s_k}$: 
% there exist an exact ODE solution form~\cite{dpmsolver} between start point $\hat{x}_{s_k}$ and intermediate point $x_t$ at any stage:
% \begin{equation}
% \label{equa:dpmsolver}
%     x_t = \frac{\alpha_t}{\alpha_{s_k}} x_{s_k} - \alpha_t \int_{\lambda_{s_k}}^{\lambda_t} e^{-\lambda} \epsilon (x_{t_\lambda}, t_\lambda) d\lambda
% \end{equation}
% where $\lambda_t=ln\frac{\alpha_t}{\beta_t}$, 
\begin{equation}
\label{equa:dpmsolver}
    x_t = \frac{\gamma_t}{\gamma_{s_k}} \hat{x}_{s_k} - \gamma_t \int_{\lambda_{s_k}}^{\lambda_t} e^{-\lambda} \epsilon (x_{t_\lambda}, t_\lambda) d\lambda,
\end{equation}
where $e_k<t<s_k$, $\lambda_t=ln\frac{\gamma_t}{\sigma_t}$, 
and $t_\lambda$ is the inverse function of $\lambda_t$. Equation~\ref{equa:dpmsolver} consists of two components: a deterministic scaling factor, given by 
% $\frac{\alpha_{t}}{\alpha_{s_k}}$, 
$\frac{\gamma_{t}}{\gamma_{s_k}}$, 
and the exponentially weighted integral of the noise $\epsilon (x_{t_\lambda}, t_\lambda)$. If $\epsilon (x_{t_\lambda}, t_\lambda)$ is a constant in stage $k$, denoted as $\epsilon_k$, the above integral is equivalent to:
% \begin{equation}
% \begin{aligned}
% \label{equa:constantsolver}
%     x_{t} = \frac{\alpha_{t}}{\alpha_{s_k}} \hat{x}_{s_k} - \alpha_{t}  \epsilon(\hat{x}_{s_k}, {s_k}) \int_{\lambda_{s_k}}^{\lambda_{t}} e^{-\lambda} d\lambda \\= \frac{\alpha_{t}}{\alpha_{s_k}} \hat{x}_{s_k} + \alpha_{t} \epsilon(\hat{x}_{s_k}, {s_k}) \left( \frac{\sigma_{t}}{\alpha_{t}} - \frac{\sigma_{s_k}}{\alpha_{s_k}} \right)
% \end{aligned}
% \end{equation}
\begin{equation}
\begin{aligned}
\label{equa:constantsolver}
    % x_{t} &= \frac{\gamma_{t}}{\gamma_{s_k}} \hat{x}_{s_k} - \gamma_{t}  \epsilon(\hat{x}_{t_\lambda}, {t_\lambda}) \int_{\lambda_{s_k}}^{\lambda_{t}} e^{-\lambda} d\lambda \\&= \frac{\gamma_{t}}{\gamma_{s_k}} \hat{x}_{s_k} + \gamma_{t} \epsilon(\hat{x}_{t_\lambda}, {t_\lambda}) \left( \frac{\sigma_{t}}{\gamma_{t}} - \frac{\sigma_{s_k}}{\gamma_{s_k}} \right).
    x_{t} &= \frac{\gamma_{t}}{\gamma_{s_k}} \hat{x}_{s_k} - \gamma_{t}  \epsilon_k \int_{\lambda_{s_k}}^{\lambda_{t}} e^{-\lambda} d\lambda \\&= \frac{\gamma_{t}}{\gamma_{s_k}} \hat{x}_{s_k} + \gamma_{t} \epsilon_k \left( \frac{\sigma_{t}}{\gamma_{t}} - \frac{\sigma_{s_k}}{\gamma_{s_k}} \right).
\end{aligned}
\end{equation}
While enforcing a constant value for $\epsilon _k$ at any stage is challenging, we can leverage data-noise alignment~\cite{immisciblediffusion} to constrain its value within a narrow range. In detail, before adding noise to video, we pre-determine the target noise distribution for each video by minimizing the aggregate distance between video-noise pairs as shown in Fig.~\ref{fig:data_noise_align}, thereby ensuring data-noise alignment and Equation~\ref{equa:dpmsolver} are approximately equivalent to Equation~\ref{equa:constantsolver}. The alignment process can be implemented using Scipy~\cite{scipy} in one line of code as shown in Algorithm~\ref{alg:batch_noise_assignment}.
\begin{algorithm}[H]
\caption{Data-Noise Alignment}
\label{alg:batch_noise_assignment}
\begin{algorithmic}[1]
% \INPUT Video batch $x$, random noise $\epsilon$
% \STATE \text{assign\_mat} $\gets$ \texttt{scipy.optimize.}
% \par \texttt{linear\_sum\_assignment}(\text{dist}(x, $\epsilon$))
% % \STATE $x_{t,b} \gets \sqrt{\alpha_{t_b}} x_b + \sqrt{1 - \alpha_{t_b}} \cdot n_{\text{rand},b}[\text{assign\_mat}]$
% % \ENSURE Diffused image batch $x_{t,b}$
% \STATE $\epsilon'\gets \epsilon$[assign\_mat]
% \OUTPUT $\epsilon'$
\REQUIRE Video batch $x$, random noise $\epsilon$
\STATE \text{assign\_mat} $\gets$ \texttt{scipy.optimize.}
\par \texttt{linear\_sum\_assignment}(\text{dist}(x, $\epsilon$))
% \STATE $x_{t,b} \gets \sqrt{\alpha_{t_b}} x_b + \sqrt{1 - \alpha_{t_b}} \cdot n_{\text{rand},b}[\text{assign\_mat}]$
% \ENSURE Diffused image batch $x_{t,b}$
\STATE $\epsilon'\gets \epsilon$[assign\_mat]
\item[\textbf{Output:}] $\epsilon'$
% \OUTPUT $\epsilon'$
\end{algorithmic}
\end{algorithm}

Our experiments demonstrate that this approximation is valid and does not compromise the model's performance.

Through data-noise alignment, we can apply Equation~\ref{equa:constantsolver} to any point in the stage, including the end point $\hat{x}_{e_k}$. 
% After obtain $\hat{x}_{e_k}$ using Equation~\ref{equa:x_e_k}, 
% substituting $\hat{x}_{e_k}$ into Equation~\ref{equa:constantsolver}, we arrive at the expression for the noise $\epsilon_k$ of stage $k$:
By substituting $t=e_k$ and $x_t=\hat{x}_{e_k}$ into Equation~\ref{equa:constantsolver}, through simple transformation, we arrive at the expression for noise $\epsilon_k$ of stage $k$ :
% the diffusion ODE sub-trajectory connecting $\hat{x}_{e_k}$ and $\hat{x}_{s_k}$:
\begin{equation}
\label{equa:epsilon}
    \epsilon_k = \frac{\frac{\hat{x}_{e_k}}{\gamma_{e_k}} - \frac{\hat{x}_{s_k}}{\gamma_{s_k}}}{\frac{\sigma_{e_k}}{\gamma_{e_k}} - \frac{\sigma_{s_k}}{\gamma_{s_k}}}.
\end{equation}
Then we can easily get any intermediate point $x_t$ in stage $k$ by substituting $\epsilon_k$ into Equation~\ref{equa:constantsolver}. Consequently, we can compute the corresponding loss using $x_t$ and $\epsilon_k$ obtained in our method and optimize model parameters in the same way as vanilla diffusion training. Note that the above derivation does not constrain the expressions of $\gamma_t$ and $\sigma_t$, making our method applicable to different diffusion frameworks. 
We also note that the direction of $\epsilon_k$ points towards the end point of the current stage rather than the final target in vanilla diffusion models as shown in 
Fig.~\ref{fig:pipeline}.
% Fig.~\ref{fig:StageDiffusion}. 
By reducing the distance between intermediate points and their target points, our method facilitates the training process and further accelerates model convergence.
% By reducing the distance to the target sample, our method effectively accelerates the training process. 
% \input{fig/stage_diffusion}

\subsection{Practical implementation}
\label{sec:practical_implementation}
In practice, 
for diffusion framework whose ODE path is curved, like DDIM, we can substitute $\gamma_t=\sqrt{\bar{\alpha_t}}$ and $\sigma_t=\sqrt{1-\bar{\alpha_t}}$ into Equation~\ref{equa:constantsolver} and Equation~\ref{equa:epsilon} to obtain $x_t$ and $\epsilon_k$. 
For flow matching, since it can transport any prior distribution to other distributions, we can model each stage as a complete flow matching process~\cite{PyramidFlow}, resulting in a simpler expression:
\begin{equation}
    x_t = (1-t')\hat{x}_{e_k}+t'\hat{x}_{s_k},
\end{equation}
where $t'=\frac{t-e_k}{s_k-e_k}$. And The objective of stage $k$ is:
\begin{equation}
    % \frac{dx_t}{dt} = \frac{\hat{x}_{s_k} - \hat{x}_{e_k}}{s_k-e_k}.
    \frac{dx_t}{dt'} = \hat{x}_{s_k} - \hat{x}_{e_k}
\end{equation} 
One aspect pyramid flow overlooks is the noise-data aligment, leading to increased variance in the prior distribution, thereby hindering model convergence. Notably, if we model each stage as a complete DDIM process, the model fails to converge. This is because it is exceedingly challenging for a single model to fit multiple curved ODE trajectories. 
% We obtain $\hat{x}_{s_k}$ and $\hat{x}_{e_k}$ for both DDIM and flow matching by:
% \begin{equation}
% \label{equa:x_s_k}
%     % \hat{x}_{s_k}=\alpha_{s_k}Up(Down(x_0,2^{k+1}))+\sigma_{s_k}\epsilon
%     \hat{x}_{s_k}=\gamma_{s_k}Up(Down(x_0,2^{k+1}))+\sigma_{s_k}\epsilon
% \end{equation}
% \begin{equation}
% \label{equa:x_e_k}
%     % \hat{x}_{e_k}=\alpha_{e_k}Down(x_0,2^k)+\sigma_{e_k}\epsilon
%     \hat{x}_{e_k}=\gamma_{e_k}Down(x_0,2^k)+\sigma_{e_k}\epsilon
% \end{equation}
% where $\epsilon\sim N(0,I)$, $Down(\cdot,2^k)$ and $Up(\cdot,2^k)$ are downsampling and upsampling $2^k$ times along temporal axis.

% \paragraph{Skip Positional Encoding}
% Video diffusion models always employ sequential positional encoding for each frame. However, since we incorporates multiple frame rates, sequential positional encoding will lead to frame mismatch between different stages as shown in Fig.~\ref{fig:skippe}. To solve this problem, we propose skip positional encoding. For the $n^{th}$ frame in stage $k$, its positional encoding $p_{n}^k$ is:
% \begin{equation}
%     p_{n}^k=PE(n*2^k)
% \end{equation}
% where PE is the positional encoding method like RoPE~\cite{rope}.

In conclusion, we visualize the training process of our method in Algorithm~\ref{alg:train}.

\begin{algorithm}[H]
\caption{Stage-wise Diffusion}
\label{alg:train}
\begin{algorithmic}[1] 
\REQUIRE Training dataset $D$, Number of stages $K$, Diffusion type DDIM or Flow Matching, Model $\epsilon_\theta$ or $v_\theta$,
Create $K$ stages $\{[s_k,e_k)\}_{k=1}^K$
\REPEAT
    \STATE Sample $x_0\sim D$;
    \STATE Sample stage $k$ from $\{1,...K\}$, then sample timestep $t\in[s_k,e_k)$
    \STATE Sample noise $\epsilon'\in N(0,I)$ aligned with $x_0$
    % , then align with $z_0$ to get $\epsilon'$
    \STATE Add $\epsilon'$ to $x_0$ and get $\hat{x}_{e_k}$ and $\hat{x}_{s_k}$
    \IF{Flow Matching}
        \STATE $x_t = (1-t')\hat{x}_{e_k}+t'\hat{x}_{s_k}$ where $t'=\frac{t-e_k}{s_k-e_k}$
        \STATE $v_k= \hat{x}_{s_k} - \hat{x}_{e_k}$
        \STATE Compute loss: $\ell=||v_\theta(x_t)-v||^2$
    \ELSE
        \STATE $\epsilon_k = \frac{\frac{\hat{x}_{e_k}}{\alpha_{e_k}} - \frac{\hat{x}_{s_k}}{\alpha_{s_k}}}{\frac{\sigma_{e_k}}{\alpha_{e_k}} - \frac{\sigma_{s_k}}{\alpha_{s_k}}}$
        \STATE $x_t = \frac{\alpha_{t}}{\alpha_{s_k}} \hat{x}_{s_k} + \alpha_{t} \epsilon_k \left( \frac{\sigma_{t}}{\alpha_{t}} - \frac{\sigma_{s_k}}{\alpha_{s_k}} \right)$
        \STATE Compute loss: $\ell=||\epsilon_\theta(x_t)-\epsilon_k||^2$
    \ENDIF
    \STATE Update $\theta$ with gradient-based optimizer using $\nabla_\theta \ell$
\UNTIL Convergence

\end{algorithmic}
\end{algorithm}

\subsection{Inference Strategy}
\label{sec:inference}
% After training, we can apply standard sampling algorithm to solve the reverse ODE. But we must carefully handle jump points~\cite{trans} between successive stages to reduce the train-inference gap. 
After training, we can use standard sampling algorithm~\cite{dpmsolver} to solve the reverse ODE in every stage. 
% However, we need to carefully handle stage continuity.
However, carefully handling stage continuity is necessary.

Upon completion of a stage, we first upsample $e_k$ in temporal dimension to double its frame rate via interpolation. Subsequently, we scale $Up(e_k)$ and inject additional random noise to match the distribution of $\hat{x}_{s_{k-1}}$ during training:
\begin{equation}
\label{equa:inf}
    \hat{x}_{s_{k-1}} = \frac{\gamma_{s_k}}{\gamma_{e_k}}Up(\hat{x}_{e_k})+\delta n', n'\sim N(0,\Sigma').
\end{equation}
Scaling factor $\frac{\gamma_{s_k}}{\gamma_{e_k}}$ ensures mean continuity while $\delta$ is utilized to compensate for the discrepancy in variance~\cite{PyramidFlow}. Considering the simplest scenario using nearest temporal upsampling and lowering the effect of noise, we derive Equation~\ref{equa:inf} as (see Appendix~\ref{appendix:inference} for detailed derivations):
\begin{equation}
\label{equa:inference}
\begin{aligned}
    \hat{x}_{s_{k-1}} = \frac{\sqrt{2}\gamma_{s_k}}{\sigma_{s_k}+\sqrt{2}\gamma_{s_k}}Up(\hat{x}_{e_k})+\frac{\sqrt{2}\sigma_t}{2}n',
\end{aligned}
\end{equation}
where,
\begin{equation}
    \\n'\sim N(0,\begin{bmatrix}
1 & -1 \\
-1 & 1 \\
\end{bmatrix}).
\end{equation}

\section{Experiments}
\label{sec:exp}
\subsection{Experimental setting}

We implement our method in both DDIM and flow matching. Since most video diffusion models are bulit upon pretrained image models, our experiments are based on two image models: MiniFlux~\cite{PyramidFlow} and SD1.5~\cite{LDM}. These two models are trained under flow matching and DDIM, respectively. We extend MiniFlux to MiniFlux-vid by finetuning all its parameters on video data and we adopt AnimateDiff~\cite{animatediff} to extend SD1.5 to video model. 
% We construct our dataset by selecting approximately 100k high-quality text-video pairs from OpenVID1M~\cite{openvid1m}. This dataset comprises videos with both motion and aesthetic scores in the top 20\%, or at least one of the scores in the top 3\%. The resolution for MiniFlux-vid and AnimateDiff is 384p and 256x256.
The number of stages is set to 3 and each stage is uniformly partitioned in all experiments. Our experiments are conducted on NVIDIA H100 GPU.

\paragraph{Dataset} We construct our dataset by selecting approximately 100k high-quality text-video pairs from OpenVID1M~\cite{openvid1m}. This dataset comprises videos with both motion and aesthetic scores in the top 20\%, or at least one of the scores in the top 3\%. The resolution for MiniFlux-vid and AnimateDiff is 384p and 256x256.

\paragraph{Baselines} 
We compare our method with the video diffusion models trained in vanilla diffusion framework. To demonstrate our approach does not lead to performance degradation, we train two baselines: MiniFlux-vid and Animatediff under vanilla flow matching and DDIM framework, without using temporal pyramid. We train these baselines from scratch on our curated dataset using the same hyperparameters as our method. To demonstrate the effectiveness of our approach compared to existing methods, we also train modelscope~\cite{modelscope} and OpenSora~\cite{opensora} from scratch on our dataset.
% Our baseline is the vanilla video diffusion model. Specifically, we train two baselines: MiniFlux-vid and Animatediff under original flow matching and DDIM framework, without using temporal pyramid. We train our baselines from scratch on our curated dataset using the same hyperparameters as our method.

% \begin{figure*}[t]
% \begin{figure}[p]
%     \centering
%     \includegraphics[width= 0.9\textwidth]{fig/images/qualitative_result_v2.pdf}
%     \caption{\textit{\textbf{Qualitative result}}. The first three rows present the results of flow matching (MiniFlux). The bottom three rows show the results of DDIM (AnimateDiff).}
%     \label{fig:qualitative_result}
% \end{figure}

% \afterpage{\clearpage}
\FloatBarrier
\begin{figure*}[t]
    \centering
    \includegraphics[width=0.9\textwidth]{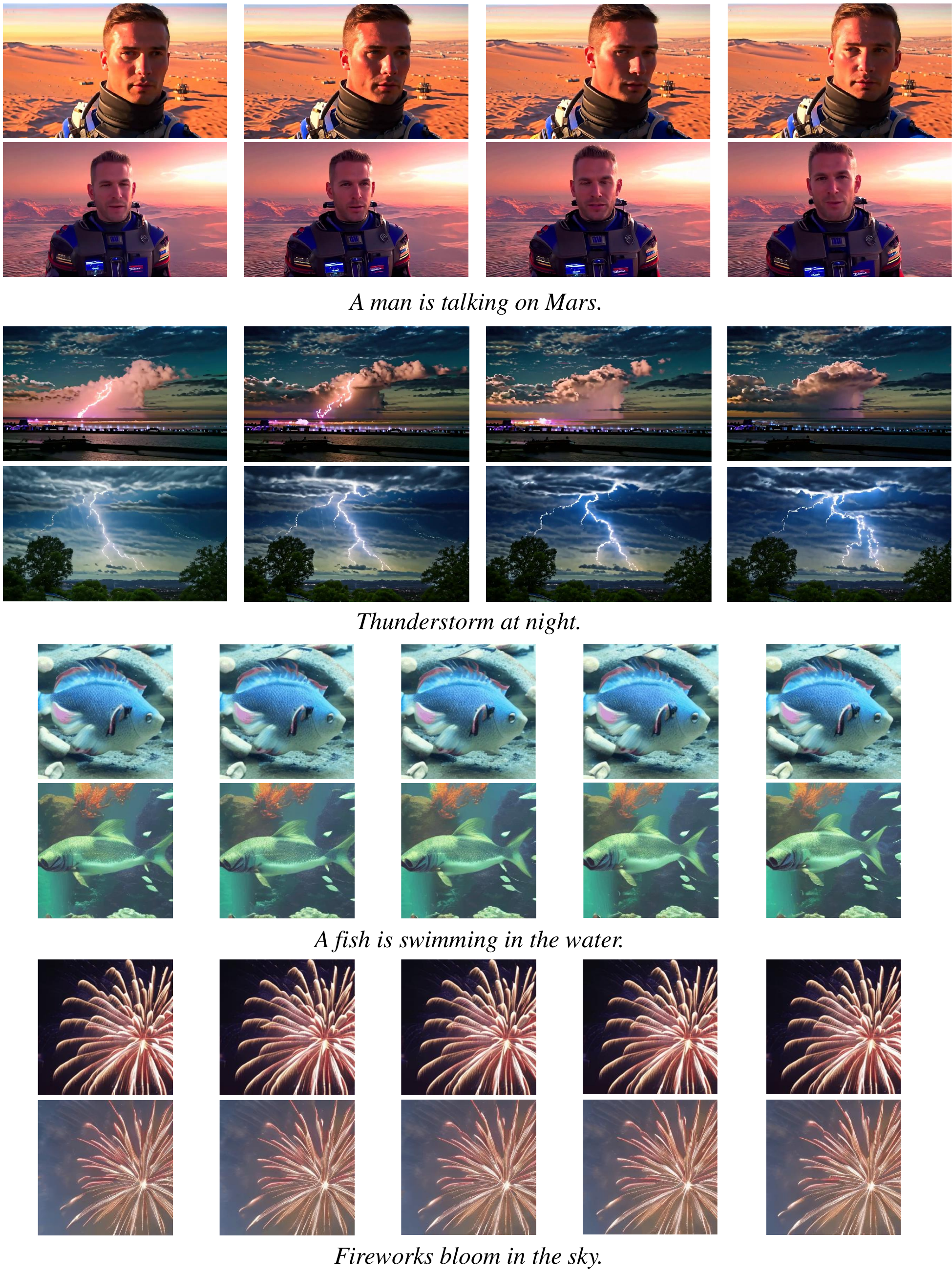}
    % \vspace{-1em}
    \caption{\textit{\textbf{Qualitative comparison}}. 
    % The first three rows present the results of flow matching (MiniFlux-vid). The bottom three rows show the results of DDIM (AnimateDiff).}
    In each pair of videos, the first row presents the results of models trained using vanilla diffusion and the second row shows the results of our method. The first two video pairs are generated by MiniFlux-vid and the remaining are generated by animatediff.}
    % \vspace{-10em}
    \label{fig:qualitative_result}
\end{figure*}
\FloatBarrier
% \afterpage{\clearpage} 

\paragraph{Evaluation} 
We evaluate our model from two perspectives: generation quality and efficiency. 
To evaluate the generation quality, we adopt quantitative metrics from
VBench~\cite{vbench} to compare our method's performance with existing models. 
For efficiency, we visualize the convergence curve to intuitively demonstrate training efficiency. In detail, to evaluate the model's generative capability during training, we follow common practice~\cite{modelscope} to use validation videos from MSRVTT~\cite{msrvtt} for zero-shot generation evaluation. We systematically compute the FVD~\cite{fvd} value during training and present the FVD-GPU hours curve to demonstrate the training efficiency of our method. 
We also report the average inference time to validate the inference efficiency.
% We also report the average time required for inference to show inference efficiency.
% For video quality, we follow common practice~\cite{modelscope} to use validation videos from MSRVTT~\cite{msrvtt} for zero-shot generation evaluation. We report standard metrics of FVD and CLIPSIM~\cite{clip}. For efficiency, we visualize the model's convergence curve to intuitively demonstrate training efficiency. We also report the average time required for inference under the same inference step.

% \input{fig/qualitative_result}

% \paragraph{Baselines} 
% We compare our method with the video diffusion models trained in vanilla diffusion framework. To demonstrate our approach does not lead to performance degradation, we train two baselines: MiniFlux-vid and Animatediff under vanilla flow matching and DDIM framework, without using temporal pyramid. We train these baselines from scratch on our curated dataset using the same hyperparameters as our method. To demonstrate the effectiveness of our approach compared to existing methods, we also train modelscope~\cite{modelscope} and OpenSora~\cite{opensora} from scratch on our dataset.
% Our baseline is the vanilla video diffusion model. Specifically, we train two baselines: MiniFlux-vid and Animatediff under original flow matching and DDIM framework, without using temporal pyramid. We train our baselines from scratch on our curated dataset using the same hyperparameters as our method.

\subsection{Quantitative results}
% \input{table/baseline_quality}

% \begin{table}[b]
% \centering
% \resizebox{\linewidth}{!}{
% \begin{tabular}{@{}l@{\hspace{2mm}}c@{\hspace{2mm}}*{3}{c@{\hspace{2mm}}}r@{}}
% % \begin{tabular}{@{}l@{\hspace{5mm}}c@{\hspace{1.5mm}}c@{\hspace{5mm}}c@{\hspace{4mm}}c@{\hspace{4mm}}c@{\hspace{3mm}}c@{\hspace{1mm}}c@{\hspace{1mm}}c@{\hspace{1mm}}c@{\hspace{1mm}}}

% \toprule
%   & FVD $\downarrow$ & CLIP-score $\uparrow$ & Cond. Recon. $\uparrow$\\
% \midrule
% X-Adapter & 30.95 & 0.2632 & 0.27 ± 0.13  \\
% X-Adapter + X-PlugVid & \textbf{30.89} & \textbf{0.2643} & \textbf{0.32 ± 0.11} \\
% \midrule
% X-Adapter & 30.95 & 0.2632 & 0.27 ± 0.13  \\
% X-Adapter + X-PlugVid & \textbf{30.89} & \textbf{0.2643} & \textbf{0.32 ± 0.11} \\
% \bottomrule
% \end{tabular}
% }
% \vspace{0.5em}
% \caption{Quantitative evaluation against X-Adapter. }
% \label{table:Quantitative_baseline}
% \end{table}

\begin{table}[h]
\centering

\vspace{1em}
\begin{tabular}{lccccccc}
\toprule
Model & Method & Latency(s)$\downarrow$ & Speed up \\ 
\midrule
\multirow{2}{*}{MiniFlux-vid} 
 & Vanilla        & 20.79 & - \\
 & \textbf{Ours}   & \textbf{12.18} & 1.71x \\
\midrule
\multirow{2}{*}{AnimateDiff} 
 & Vanilla        & 6.01 & - \\
 & \textbf{Ours}   & \textbf{4.04} & 1.49x \\
\bottomrule
\end{tabular}
\caption{Inference efficiency of baselines and our method. The total denoising step is set to 30 for all models.}
\label{tab:baseline_efficiency}
\end{table}

Tab.~\ref{tab:baseline_quality} shows quantitative comparison between our method and baselines. Compared to existing method, our model achieves better results with a higher total score. Compared to the vanilla diffusion models, our approach demonstrates improvements in most aspects, demonstrating that it enhances efficiency without compromising performance.
% while further facilitating more effective modeling of spatial and temporal relationships. 
This further indicates that the vanilla video diffusion model contains substantial redundancy in temporal modeling, whereas our approach effectively eliminates such redundancies. 

\begin{figure}[t]
    \centering
    \includegraphics[width=0.8\columnwidth]{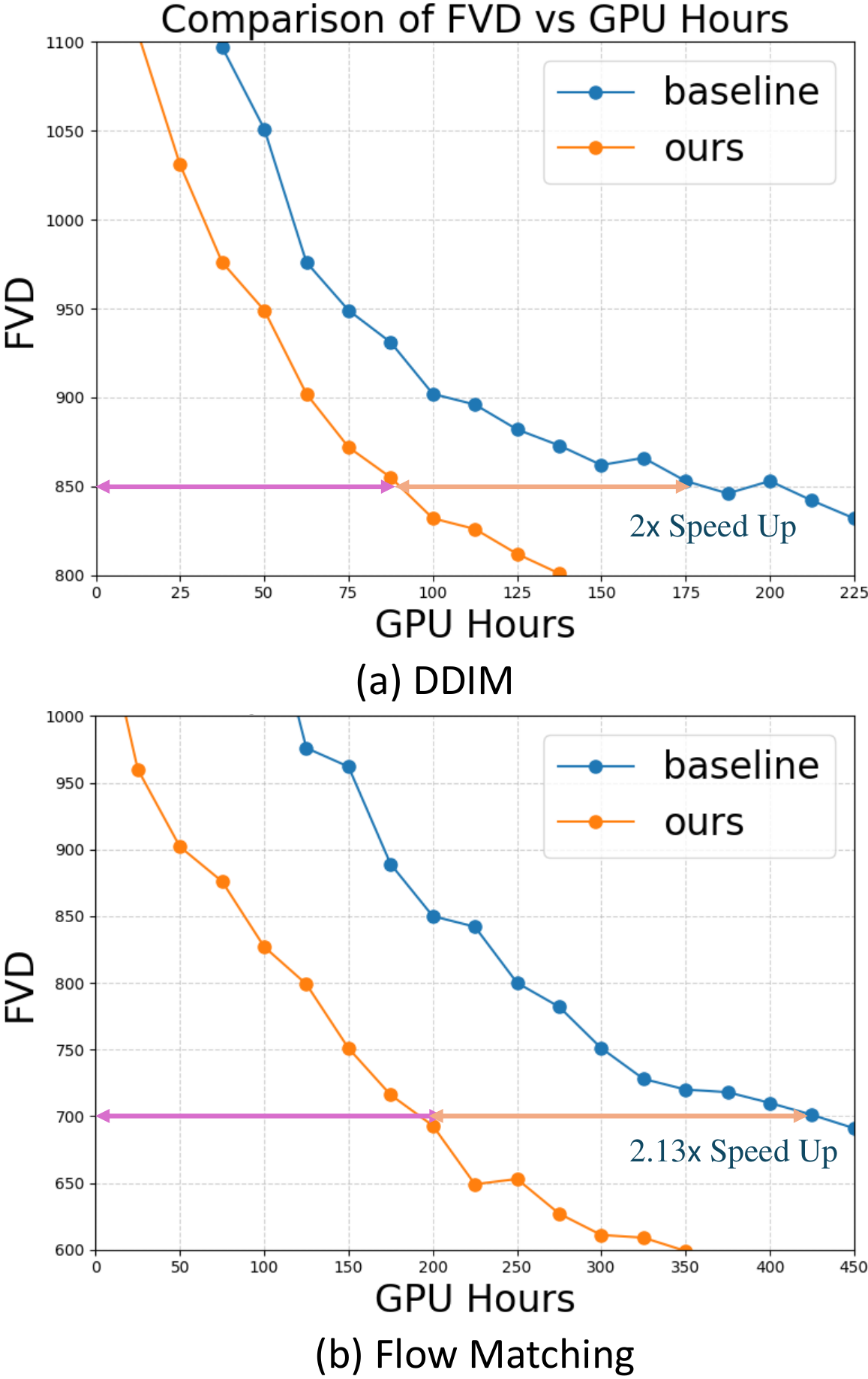}
    % \vspace{-1em}
    \caption{\textit{\textbf{Convergence curve of vanilla diffusion models and our method on (a) DDIM, (b) Flow Matching}}. We illustrate the FVD of two methods with different GPU hours consumed. Our method achieves higher training efficiency compared to vanilla approachs.}
    \label{fig:converge_curve}
\end{figure}

Figure.~\ref{fig:converge_curve} shows that our method achieves speedup of 2x and 2.13x in training compared to vanilla diffusion models. This acceleration primarily stems from two factors: 
1) Noise-data pairing: By aligning noise with data, we reduce the randomness in training. The model learns a nearly deterministic ODE path rather than the expectation of multiple intersecting ODE paths. 
2) Shorter averaged sequence length. Since the computational complexity of attention mechanism scales quadratically with sequence length, our method requires significantly less computational complexity on average. For example, to generate a video of length $T$, the averaged computational cost of attention modules in our method is halved, reducing to $\frac{1}{3}(T^2+(\frac{T}{2})^2+(\frac{T}{4})^2)\approx 0.44T^2$ compared to $T^2$ in vanilla diffusion model. This advantage is also reflected in faster inference speed as shown in Tab.~\ref{tab:baseline_efficiency}.

% \begin{table}[h]
% \label{tab:baseline_quality}
% \centering
% \caption{Video generaration quality of baselines and our method.}
% \vspace{1em}
% \begin{tabular}{lccccccc}
% \toprule
% \multirow{2}{*}{Model} & \multirow{2}{*}{Method} & \multicolumn{2}{c}{Metrics} \\ 
% \cmidrule(lr){3-4} 
%  &  & FVD$\downarrow$ & CLIPSIM$\uparrow$ \\
% \midrule
% \multirow{2}{*}{MiniFlux-vid} 
%  & Original        & 592.1 & 29.43 \\
%  & \textbf{Ours}   & \textbf{541.6} & \textbf{29.83} \\
% \midrule
% \multirow{2}{*}{AnimateDiff} 
%  & Original        & 839.2 & 28.12 \\
%  & \textbf{Ours}   & \textbf{774.5} & \textbf{28.89} \\
% \bottomrule
% \end{tabular}
% \end{table}

% \begin{table*}[h]
% \label{tab:baseline_quality}
%     \centering
%     \begin{tabular}{lccccc}
%         \toprule
%         Method & Total Score & Motion Smoothness & Object Class & Multiple Objects & Spatial Relationship \\
%         \midrule
%         a & -\% & -\% & -\% & -\% & -\% \\
%         b & -\% & -\% & -\% & -\% & -\% \\
%         c & -\% & -\% & -\% & -\% & -\% \\
%         d & -\% & -\% & -\% & -\% & -\% \\
%         \midrule
%         \textbf{\lm{method name} (Ours)} & \textbf{-\%} & \textbf{-\%} & \textbf{-\%} & \textbf{-\%} & \textbf{-\%} \\
%         \textbf{\lm{method name} (Ours)} & \textbf{-\%} & \textbf{-\%} & \textbf{-\%} & \textbf{-\%} & \textbf{-\%} \\
%         \bottomrule
%     \end{tabular}
%     \caption{Comparison of video generation quality of baselines and our method.}
%     \label{tab:results}
% \end{table*}

\begin{table*}[h]
\centering
    \begin{tabular}{lccccc}
        \toprule
        Method & Total Score & Motion Smoothness & Object Class & Multiple Objects & Spatial Relationship \\
        \midrule
        ModelScope~\cite{modelscope} & 73.12\% & 95.83\% & 67.06\% & 33.91\% & 27.55\% \\
        OpenSora~\cite{opensora} & 74.08\% & 91.97\% & 78.30\% & 30.69\% & 41.11\% \\
        % \midrule
        % \textbf{\lm{method name} (Ours)} & \textbf{-\%} & \textbf{-\%} & \textbf{-\%} & \textbf{-\%} & \textbf{-\%} \\
        % \textbf{\lm{method name} (Ours)} & \textbf{-\%} & \textbf{-\%} & \textbf{-\%} & \textbf{-\%} & \textbf{-\%} \\
        \midrule
        % \multirow{2}{*}{%
        %   \begin{tabular}{@{}cc@{}}
        %     AnimateDiff~\cite{animatediff} & Vanilla
        %     \\ & Ours
        %   \end{tabular}%
        % }
        % & 73.65\% & 96.28\% & 81.34\% & \textbf{\textcolor{blue}{39.42\%}} & 43.23\% \\
        % & \textbf{\textcolor{blue}{74.96\%}} & \textbf{\textcolor{blue}{97.68\%}} & \textbf{\textcolor{blue}{86.77\%}} & 31.15\% & \textbf{\textcolor{blue}{56.62\%}} \\
        AnimateDiff - Vanilla
        & 73.65\% & 96.28\% & 81.34\% & \textbf{\textcolor{blue}{39.42\%}} & 43.23\% \\
        AnimateDiff - Ours
        & \textbf{\textcolor{blue}{74.96\%}} & \textbf{\textcolor{blue}{97.68\%}} & \textbf{\textcolor{blue}{86.77\%}} & 31.15\% & \textbf{\textcolor{blue}{56.62\%}} \\
        \midrule
        % \multirow{2}{*}{%
        %   \begin{tabular}{@{}cc@{}}
        %     MiniFlux-vid~\cite{PyramidFlow} & Vanilla
        %     \\ & Ours
        %   \end{tabular}%
        % }
        MiniFlux-vid - Vanilla & 77.69\% & 98.34\% & 77.32\% & 36.26\% & \textbf{49.79\%}\\
        MiniFlux-vid - Ours & \textbf{78.52\%} & \textbf{98.96\%} & \textbf{83.21\%} & \textbf{43.18\%} & 42.23\% \\
        \bottomrule
    \end{tabular}
    \caption{Comparison of video generation quality of baselines and our method.}
    \vspace{-1em}
    \label{tab:baseline_quality}
\end{table*}

% \begin{table*}[h]
% \centering
%     \begin{tabular}{lccccc}
%         \toprule
%         Method & Total Score & Motion Smoothness & Object Class & Multiple Objects & Spatial Relationship \\
%         \midrule
%         ModelScope~\cite{modelscope} & 73.12\% & 95.83\% & 67.06\% & 33.91\% & 27.55\% \\
%         OpenSora~\cite{opensora} & 74.08\% & 91.97\% & 78.30\% & 30.69\% & 41.11\% \\
%         \midrule
%         \multirow{2}{*}{AnimateDiff~\cite{animatediff}} & Vanilla & 73.65\% & 96.28\% & 81.34\% & \textbf{\textcolor{blue}{39.42\%}} & 43.23\% \\
%          & Ours & \textbf{\textcolor{blue}{74.96\%}} & \textbf{\textcolor{blue}{97.68\%}} & \textbf{\textcolor{blue}{86.77\%}} & 31.15\% & \textbf{\textcolor{blue}{56.62\%}} \\
%         \midrule
%         \multirow{2}{*}{MiniFlux-vid~\cite{PyramidFlow}} & Vanilla & 77.69\% & \textbf{98.94}\% & 77.32\% & 36.26\% & 42.23\% \\
%          & Ours & \textbf{78.52\%} & 98.36\% & \textbf{83.21\%} & \textbf{43.18\%} & \textbf{49.79\%} \\
%         \bottomrule
%     \end{tabular}
%     \caption{Comparison of video generation quality of baselines and our method.}
%     \label{tab:baseline_quality}
% \end{table*}

\begin{figure}[t]
    \centering
    \includegraphics[width=\columnwidth]{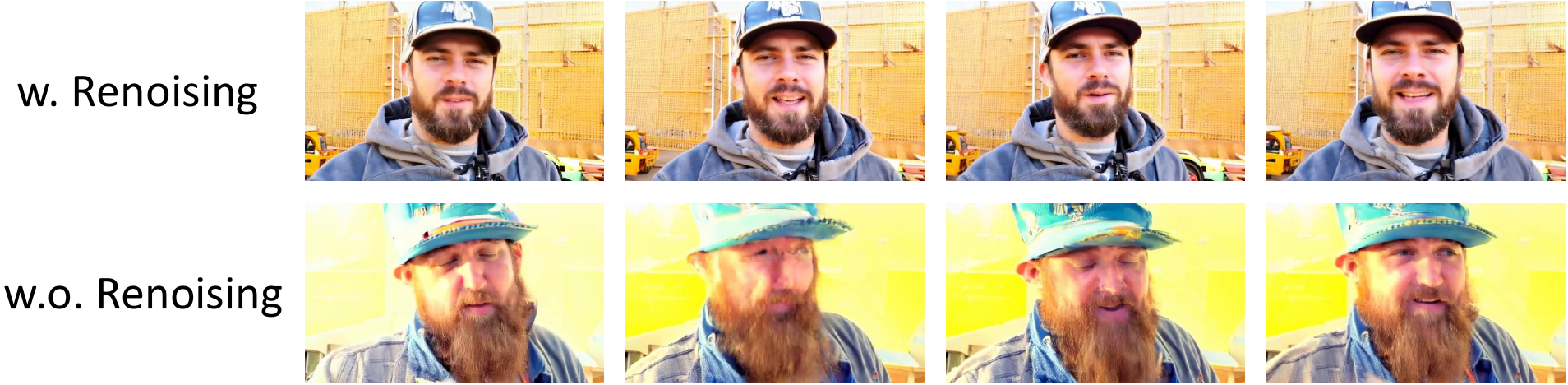}
    \caption{\textit{\textbf{Ablation study of inference strategy.}} Our method generates smooth, high-quality videos, whereas the baseline without inference renoising exhibits significant flickers}
    \label{fig:ablation_inference}
\end{figure}

\begin{figure}[t]
    \centering
    \includegraphics[width=\columnwidth]{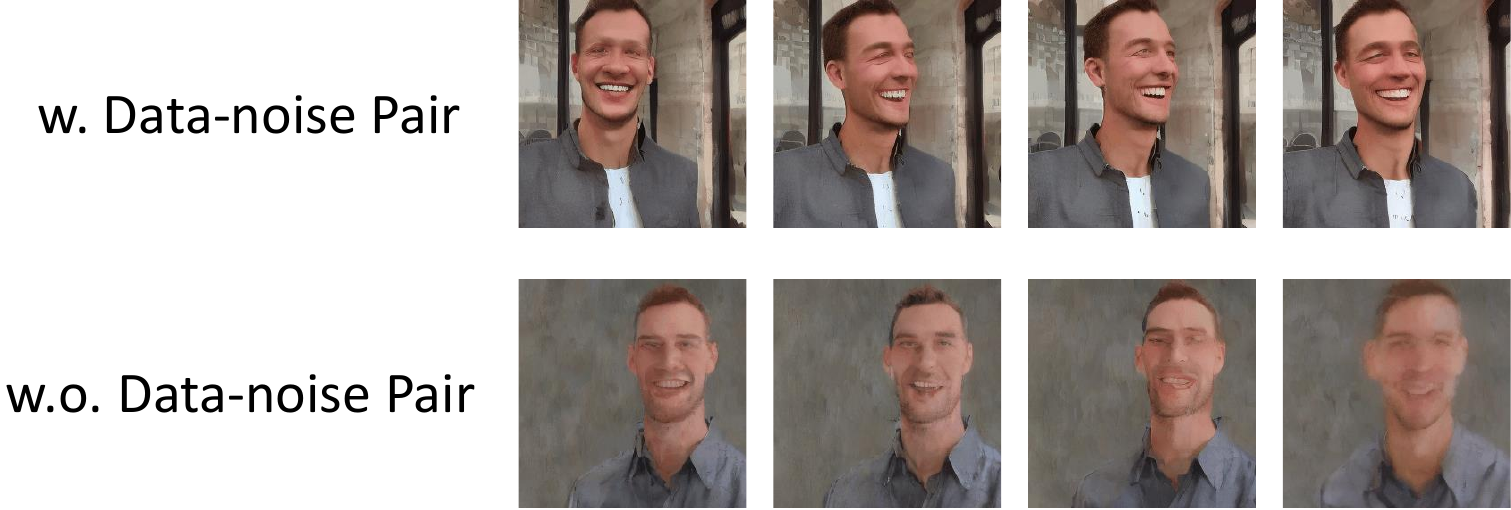}
    \caption{\textit{\textbf{Ablation study of data-noise alignment.}} Our method can produce clearer videos compared to the baseline.}
    \label{fig:ablation_data_noise_pair}
\end{figure}

\subsection{Qualitative result}
As shown in Fig.~\ref{fig:qualitative_result}, we show qualitative comparison between our method and vanilla video diffusion models. The results generated by our method are presented in the second column and the outputs of our baseline are displayed in the first column. Evidently, our approach is able to generate videos with better semantic accuracy and larger motion. For instance, under prompt "A man is talking on Mars", the baseline generates a person merely shaking their head without speaking, failing to fully adhere to the prompt. In contrast, our approach accurately generates the specified actions, demonstrating superior alignment with the given prompt. Moreover, for AnimateDiff, the baseline generates videos that are nearly static, whereas our approach achieves motion with a more natural and reasonable amplitude.
% We provide more qualitative results in the supplementary material.

\subsection{Ablation study}

We conduct ablation study on two key designs: data-noise alignment and renoising inference strategy.

\paragraph{Ablation on data-noise alignment} 
To demonstrate the effectiveness of data-noise alignment, we curate a baseline that trains without alignment. Fig.~\ref{fig:ablation_data_noise_pair} and Tab.~\ref{tab:ablation_data_noise_alignment} present comparison between our method and this variant. Our method is capable of generating high-quality and smooth videos, whereas the baseline produces blurred results. 
This is because, without alignment, the approximation from Equation~\ref{equa:dpmsolver} to Equation~\ref{equa:constantsolver} incurs increased error. Consequently, $x_t$ and $\epsilon_k$ computed via Equation~\ref{equa:constantsolver} and Equation~\ref{equa:epsilon} deviate from the true values, leading to blurred results.
\begin{table}[h]
\centering
\begin{tabular}{cccccccc}
\toprule
\multirow{2}{*}{Model} & \multirow{2}{*}{Method} & \multicolumn{2}{c}{Metrics} \\ 
\cmidrule(lr){3-4} 
 &  & FVD$\downarrow$ & CLIPSIM$\uparrow$ \\
\midrule
\multirow{2}{*}{MiniFlux-vid} 
 & w.o. alignment        & 602.1 & 29.34 \\
 & w. alignment   & \textbf{562.6} & \textbf{29.76} \\
\midrule
\multirow{2}{*}{AnimateDiff} 
 & w.o. alignment          & 834.5 & 28.21 \\
 & w. alignment   & \textbf{782.6} & \textbf{28.91} \\
\bottomrule
\end{tabular}
\caption{Ablation on data-noise alignmenmt.}
\vspace{-1em}
\label{tab:ablation_data_noise_alignment}
\end{table}

\paragraph{Ablation on renoising inference strategy} 
We study the effect of this strategy by comparing inference with and without renoising as shown in Fig.~\ref{fig:ablation_inference}. The results indicate that while not using corrective noises can still produce generally coherent videos, it inevitably leads to flickers and blurred result.

\section{Discussion}
\label{sec:discussion}
\begin{figure}[t]
    \centering
    \includegraphics[width=\columnwidth]{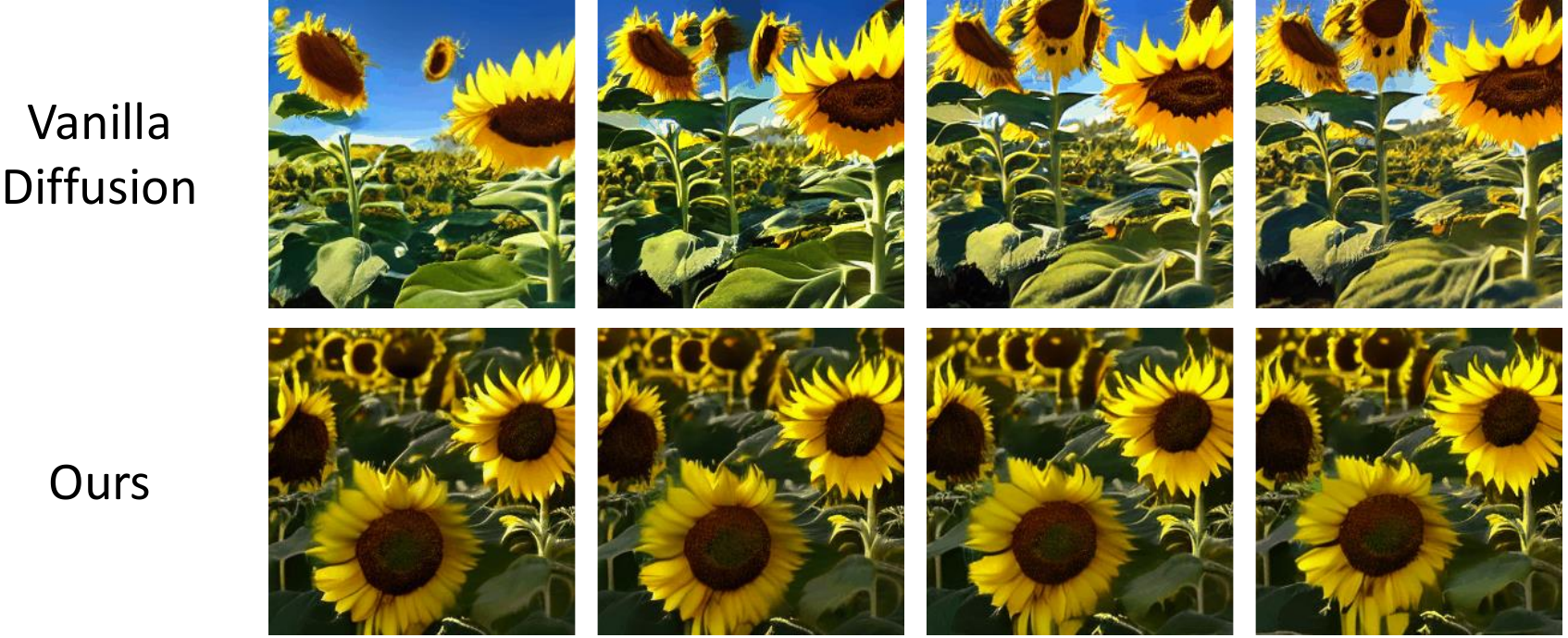}
    \caption{\textit{\textbf{Comparison between vanilla diffusion and our method after 5000 training steps.}} Our method can generate temporally stable videos even at very early training steps while vanilla method cannot. The prompt is "A serene scene of a sunflower field."}
    \label{fig:discussion}
    \vspace{-1em}
\end{figure}

In our experiments, we observe that our approach is capable of generating temporally stable videos even at very early training steps as shown in Fig.~\ref{fig:discussion}. For results of the vanilla diffusion model, the sunflower appears abruptly, whereas our method achieves much smoother camera movement. This is attributed to the temporal pyramid, which alleviates the need to learn the temporal relation of all frames under low SNR timesteps where inter-frame connections are actually absent. Consequently, our method achieves better visual quality and motion dynamics. 
\section{Conclusion}
\label{sec:conclusion}
In this paper, we propose a general acceleration framework for video diffusion models. We introduce TPDiff, a framework that progressively increases the frame rate along the diffusion process. Moreover, we design a dedicated training framework named stage-wise diffusion, which is applicable to any form of diffusion. Our experiments demonstrate that our method accelerates both training and inference on different frameworks.
{
    \small
    \bibliographystyle{ieeenat_fullname}
    \bibliography{main}
}

\newpage
\appendix
\onecolumn
\section{Appendix.}

\subsection{DERIVATION}
\label{appendix:inference}
This section provides derivation for Equation.~\ref{equa:inference}. Our derivation primarily follows pyramid flow~\cite{PyramidFlow}, and we extend it to the temporal dimension. According to Equation.~\ref{equa:x_e_k} and Equation.~\ref{equa:x_s_k}:
\begin{equation}
\begin{aligned}
    \hat{x}_{s_k} &\sim N(\gamma_{s_k}Up(Down(x_0,2^{k+1})),\sigma_{s_k}^2I) \\
    Up(\hat{x}_{e_{k+1}}) &\sim N(\gamma_{e_{k+1}}Up(Down(x_0,2^{k+1})),\sigma_{e_{k+1}}I)
\end{aligned}
\end{equation}

Spatial pyramid has demonstrate that stages can be smoothly connected by renoising the endpoint of the last stage. Renoising process can be expressed as:
\begin{equation}
\hat{x}_{s_k}=\frac{\gamma_{s_k}}{\gamma_{e_{k+1}}}Up(\hat{x}_{e_{k+1}})+\alpha n', n\sim N(0,\Sigma')
\end{equation}
where the rescaling coefficient $\frac{s_k}{e_{k+1}}$ allows the means of these distributions to be matched, and $\alpha$ is the noise weight. Additionally, we need to match the covariance matrices:
\begin{equation}
\label{equa:covariance}
    \frac{\gamma_{s_k}^2}{\gamma_{e_{k+1}}^2} \sigma_{k+1}^2 \Sigma + \alpha^2 \Sigma' =  \sigma_{s_k}^2 I.
\end{equation}
we consider the simplest interpolation: nearest neighbor temporal upsampling. Then we can get upsampling $\Sigma$ and noise's covariance matrix $\Sigma'$ has the same structure as $\Sigma$:
\begin{equation}
\label{equa:sigma}
\Sigma_{block}=\begin{pmatrix}
1 & 1 \\
1 & 1 \\
\end{pmatrix}
\implies
\Sigma'=\begin{pmatrix}
1 & \gamma \\
\gamma & 1 \\
\end{pmatrix}
\end{equation}
To ensure $\Sigma'$ is semidefinite, $\gamma \in [-1,0]$. Then we solve Equation.~\ref{equa:covariance} and Equation.~\ref{equa:sigma} by considering the equality of their diagonal and non-diagonal elements and get the solution:
\begin{equation}
\label{equa:alpha}
\begin{aligned}
\gamma_{e_{k+1}} = \frac{\gamma_{s_k} \sqrt{1 - \gamma}}{\sigma_{s_k} \sqrt{-\gamma} + \gamma_{s_k} \sqrt{1 - \gamma}}, \quad
\alpha = \frac{\sigma_{s_k}}{\sqrt{1 - \gamma}}
\end{aligned}
\end{equation}
To reduce the affect of noise, let $\gamma=-1$ and substitute it into Equation.~\ref{equa:alpha}, we can get:
\begin{equation}
\gamma_{e_{k+1}}=\frac{\sqrt{2}\gamma_{s_k}}{\sigma_{s_k}+\sqrt{2}\gamma_{s_k}},\alpha=\frac{\sqrt{2}\sigma_t}{2}
\end{equation}
We can finally obtain Equation.~\ref{equa:inference}:
\begin{equation}
\label{equa:inference}
\begin{aligned}
    \hat{x}_{s_{k-1}} = \frac{\sqrt{2}\gamma_{s_k}}{\sigma_{s_k}+\sqrt{2}\gamma_{s_k}}Up(\hat{x}_{e_k})+\frac{\sqrt{2}\sigma_t}{2}n'
\end{aligned}
\end{equation}
where:
\begin{equation}
    \\n'\sim N(0,\begin{bmatrix}
1 & -1 \\
-1 & 1 \\
\end{bmatrix})
\end{equation}

\end{document}